%% file: main.tex
\documentclass[lettersize,journal]{IEEEtran}
\usepackage{amsmath,amsfonts}
\usepackage{algorithmic}
\usepackage{algorithm}
\usepackage{array}
\usepackage[caption=false,font=normalsize,labelfont=sf,textfont=sf]{subfig}
\usepackage{textcomp}
\usepackage{stfloats}
\usepackage{url}
\usepackage{verbatim}
\usepackage{graphicx}
\usepackage{cite}
\usepackage{multicol}
\usepackage{multirow}

\usepackage{booktabs}
\usepackage{makecell}
\usepackage{pifont}
\usepackage{colortbl}
\usepackage{xcolor}
\usepackage{array}
\usepackage{verbatim}

\definecolor{MyDarkRed}{rgb}{0.8,0.02,0.02}

\usepackage{CJKutf8}

\usepackage{hyperref}
\hypersetup{
    colorlinks=true,
    linkcolor=blue,
    citecolor=blue,
    urlcolor=blue
}
\usepackage{cleveref}

\begin{document}

\title{Patch-Discontinuity Mining for Generalized Deepfake Detection}

\author{Huanhuan Yuan, Yang Ping\textsuperscript{*}, Zhengqin Xu,  Junyi Cao, Shuai Jia, and Chao Ma\textsuperscript{*}

\thanks{Huanhuan Yuan is with the Artificial Intelligence Institute, Shanghai Jiao Tong University, Shanghai 200240. (e-mail: yuanhuanhuan@sjtu.edu.cn)}
\thanks{Yang Ping (Co-corresponding author) is with the Chinese Academy of Military Science, Beijing 100091. (e-mail: ocean.py@163.com)}
\thanks{Zhengqin Xu, Junyi Cao, Shuai Jia, and Chao Ma (Co-corresponding author) are with the Artificial Intelligence Institute, Shanghai Jiao Tong University, Shanghai 200240. (e-mail: fate311@sjtu.edu.cn; junyicao@sjtu.edu.cn; jiashuai@sjtu.edu.cn; chaoma@sjtu.edu.cn)}
}

\maketitle

\begin{abstract}
The advancement of generative artificial intelligence has led to the creation of more diverse and realistic fake facial images. This poses serious threats to personal privacy and can contribute to the spread of misinformation. Existing deepfake detection methods usually utilize prior knowledge about forged clues to design complex modules, achieving excellent performance in the intra-domain settings. However, their performance usually suffers from a significant decline in unseen forgery patterns. It is thus desirable to develop a generalized deepfake detection method using a neat network structure. In this paper, we propose a simple yet efficient framework to transfer a powerful large-scale vision model like ViT to the downstream deepfake detection task, namely the generalized deepfake detection framework (GenDF). Concretely, we first propose a deepfake-specific representation learning (DSRL) scheme to learn different discontinuity patterns across patches inside a fake facial image and continuity between patches within a real counterpart in a low-dimensional space. To further alleviate the distribution mismatch between generic real images and human facial images consisting of both real and fake, we introduce a feature space redistribution (FSR) scheme to separately optimize the distributions of real and fake feature space, enabling the model to learn more distinctive representations. Furthermore, to enhance the generalization performance on unseen forgery patterns produced by constantly evolving facial manipulation techniques and diverse variations on real faces, we propose a classification-invariant feature augmentation (CIFAug) function without trainable parameters. CIFAug expands the scopes of real and fake feature space along directions orthogonal to the classification direction, enabling the model to learn more generalizable features while preserving discrimination. Extensive experiments demonstrate that our method achieves state-of-the-art generalization performance in cross-domain and cross-manipulation settings with only 0.28M trainable parameters. \textit{Project page: \url{https://gendf.github.io/}.}

\end{abstract}


\input{section/1_Introduction}

\input{section/2_Related_work}

\input{section/3_Method}

\input{section/4_Experiment}

\input{section/5_Conclusion}

\section*{Acknowledgments}
This work was supported in part by NSFC (62322113, 62376156), Shanghai Municipal Science and Technology Major Project (2021SHZDZX0102), and the Fundamental Research Funds for the Central Universities.

\nocite{*}
\bibliography{ref} 
\bibliographystyle{IEEEtran}

\end{document}

%% file: section/1_Introduction.tex
\section{Introduction}
\begin{figure}[t]
\centering
\includegraphics[width=1.0\linewidth]{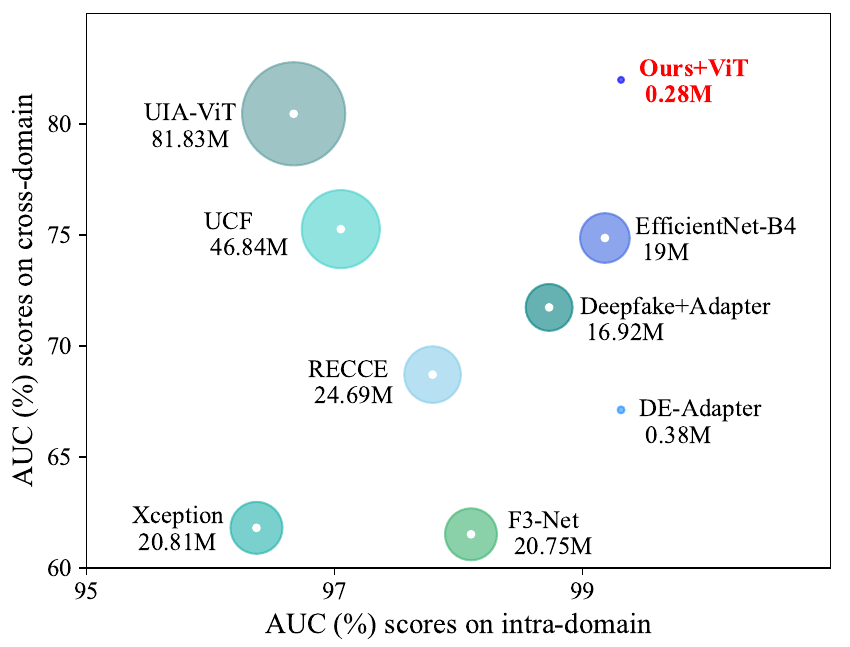}
\caption{Comparison of existing deepfake detection methods and our method in terms of the number of trainable parameters and intra-domain and cross-domain performance. The x-axis denotes the AUC results with training and testing on FF++(HQ), while the y-axis represents the AUC results with training on FF++(HQ) and testing on the Celeb-DF dataset. The size of the circles indicates the scale of trainable parameters. Our framework achieves the best generalization performance among all methods with the fewest trainable parameters.}
\label{scatter}
\end{figure}

With recent promising advances in generative artificial intelligence (AI), the visual quality of manipulated face images becomes increasingly realistic and natural. In particular, users can handily manipulate faces by utilizing off-the-shelf face editing tools, including Deepfakes~\cite{2020deepfacelab} and Faceswap~\cite{faceswap} or some faceswapping  algorithms~\cite{li2024learning,li2019faceshifter}. While manipulation techniques have been leveraged for meaningful applications like marketing and entertainment, they can also be used maliciously to create fake facial images and deceive people with fake news or political rumors, leading to growing social concerns. Thus, developing effective deepfake detection methods to counter malicious attacks is essential.

Although many studies~\cite{liu2023,yu2023,recce-cvpr2022, FF++xception-iccv2019, F3-net, FedForgery2023, Twobranch} have explored deepfake detection, the generalization capabilities of these models remain constrained by the forgery types they have encountered. Therefore, enhancing the generalization performance of face forgery detection remains a challenging problem. Recent works~\cite{LAE2020,recce-cvpr2022,global-local, RFFR2023, FedForgery2023,luo2023beyond,dong2023implicit} aim to address this challenge by developing deep learning models that incorporate prior knowledge, such as forged traces at local parts or global parts of an image. While these methods achieve impressive results, their increasingly complex structures require careful design and tend to capture method-specific patterns, leading to overfitting on the training datasets and poor generalization to unseen forgery patterns. 

In the era of large-scale models, the pretrained vision models~\cite{vit-iclr2021, 2023sam,liu2021swin} have the potential to enhance the generalization performance of deepfake detection due to their powerful capability of learning representations. As a remarkable large-scale vision model, Vision Transformer (ViT)~\cite{vit-iclr2021} reshapes an image into a sequence of 2D patches and captures the relationship between these patches. It attains excellent results on some image recognition benchmarks and possesses powerful generic image classification ability. Some recent works~\cite{uia-vit-eccv2022,de-adapter2024,deepake-adapter2023} propose fine-tuning methods to adapt the pretrained vision model for deepfake detection.

UIA-ViT~\cite{uia-vit-eccv2022} captures local forged clues in an unsupervised way while it trains all parameters of ViT, resulting in massive computational overhead and slow convergence. Some recent works leverage parameter-efficient fine-tuning (PEFT) methods for face forgery detection~\cite{vit_lora-2023,deepake-adapter2023, DADF,de-adapter2024}. ViT+LoRA~\cite{vit_lora-2023} and DE-Adapter~\cite{de-adapter2024} fine-tune large-scale vision model (LVMs) with few trainable parameters, while their generalizable performances of cross-manipulation and cross-dataset are insufficient. DeepFake-Adapter~\cite{deepake-adapter2023} and DADF~\cite{DADF} devise complex adapters, resulting in a large number of trainable parameters and low generalization on cross-domain. 
\\

To tackle the challenges mentioned above, we designed a simple yet efficient deepfake detection method to improve generalization performance in unseen forged facial image scenarios. Specifically, we propose a generalized deepfake detection framework (dubbed GenDF) based on large-scale vision models (LVMs) like ViT, driven by three considerations. Firstly, our core insight is that the fundamental distinction between real and fake facial images lies in the continuity of pixel values. Real images typically incorporate smooth and consistent pixel values, whereas fake images introduce discontinuities due to subtle pixel-level manipulations, regardless of the specific forgery method. Additionally, the Vision Transformer (ViT), with its self-attention mechanism, is inherently capable of modeling relationships between different regions of an image. In addition, previous works~\cite{r21,r22} have demonstrated that pretrained large-scale models essentially reside in a low intrinsic dimension space and can effectively learn even after being randomly projected onto a smaller subspace. Motivated by these observations, we propose a Deepfake-Specific Representations Learning (DSRL) scheme to transfer knowledge from generic real images to human facial images by fine-tuning ViT in a low-dimensional feature space with a small number of trainable parameters. DSRL helps capture the continuity between patches of real faces and the various discontinuity patterns between patches of fake faces. However, the pretrained ViT lacks basic knowledge of the domains containing real and fake human faces, as it is trained on generic real images. Consequently, fine-tuning ViT with DSRL is insufficient. To further address the distribution mismatch between generic images and real and fake facial images, we introduce the Feature Space Redistribution (FSR) transformation, which separately optimizes the distribution of real and fake feature spaces, facilitating more discriminative representations. However, real facial images usually exhibit diverse variations introduced by contrast, ethnicity, lighting, and facial angles. Furthermore, the continuous advancement of deepfake techniques constantly leads to the emergence of new types of forgeries. These factors further enhance the difficulty of distinguishing real and fake faces in unseen domains. Building upon the previous two steps, we propose the class-invariant feature augmentation (CIFAug) function. CIFAug enhances feature diversity by expanding the scopes of real and fake feature spaces along directions orthogonal to the classification direction, enabling the model to learn generalizable representations while preserving discriminability between real and fake facial images.
 
 Extensive experiments on multiple datasets illustrate that our method achieves the best generalization performance against existing methods, involving only 0.28M trainable parameters, as reported in Fig.~\ref{scatter}.

The contributions of our paper are summarized as follows:
\begin{itemize}

\item{We propose GenDF, a simple yet effective deepfake detection framework that learns diverse patch discontinuity patterns in fake facial images and patch continuity in real ones by fine-tuning ViT with a small number of trainable parameters. Our approach significantly enhances generalization performance in both cross-manipulation and cross-domain settings while also achieving faster convergence.}

\item{We develop the Deepfake-Specific Representations Learning (DSRL) scheme to capture essential differences between real and forged faces. By fine-tuning ViT with a small number of parameters, DSRL recognizes continuity between patches within a real facial image and the different discontinuity patterns across patches inside a fake counterpart. To further mitigate the distribution mismatch between generic camera-captured images and human facial images containing both real and fake types, we introduce the Feature Space Redistribution (FSR) module. FSR refines the feature distributions of real and forged faces, enhancing their separability and improving the discriminative capability of our model.}

\item{We further design the class-invariant feature augmentation (CIFAug) function to improve the generalization performance of deepfake detection. CIFAug expands real and fake feature space scopes along the directions orthogonal to the classification direction without introducing additional trainable parameters, ensuring efficiency while enhancing generalization performance on unseen scenarios.}

\end{itemize}

%% file: section/2_Related_work.tex
\section{Related Work}

\subsection{Deepfake Detection}
Existing deepfake detection methods usually formulate a binary classification task to distinguish real and fake facial images. Early methods mainly identify the noticeable counterfeit artifacts or inconsistencies with hand-crafted features, such as eye blinking~\cite{r1,r3}, inconsistent head pose~\cite{r2}, etc. However, these traditional methods are time-consuming and have become less effective due to the improvement of synthetic techniques. With the advancement of deep learning, existing methods~\cite{liu2023,yu2023, F3-net,ru2021bita} mainly develop deep neural networks to achieve better feature representation capabilities. Nicolò Bonettini\textit{et al.}~\cite{r39} utilize the EfficientNetB4~\cite{Efficientnet-icml2019} based on the CNN network to improve the efficiency of deepfake detection.
Another approach~\cite{xception} designs the pretrained Xception network and achieves superior detection performance. However, simple CNN architectures tend to suffer from severe performance degradation when encountering diverse datasets due to their limited receptive field. Therefore, recent deepfake detection methods~\cite{LAE2020,recce-cvpr2022,global-local,vit_lora-2023, RFFR2023, FedForgery2023,wang2023altfreezing} focus on improving generalization performance in unknown domains. Some works~\cite{global-local,r18} combine global and local features to learn rich representations for deepfake detection. Ju \textit{et al.}~\cite{global-local} fuse multi-scale global semantic features and detailed local artifacts from refined patches.

Another branch of deepfake detection methods focuses on the difference between real and fake data distributions. The study in~\cite{RFFR2023} learns the distribution of real facial images from large-scale datasets to obtain a general representation and detect fake facial images outside of this distribution. The work assumes a boundary of real face space exists, but the boundary could be inaccurate and susceptible to noisy data. Guo\textit{ et al.}~\cite{r19} increase the discrimination among various forgery patterns, enabling the model to concentrate more on specific forgery cues rather than the irrelevant features, such as hair color.
In addition, RECCE~\cite{recce-cvpr2022} emphasizes the real faces with reconstruction learning. The UIA-ViT~\cite{uia-vit-eccv2022} fully fine-tunes the large vision model ViT~\cite{vit-iclr2021} and learns inconsistency-aware features without pixel-level annotations, which makes the ViT model over-fitted to the small downstream dataset and causes a slow convergence rate. In conclusion, these previous methods exhibit enhanced generalization performance in cross-domain settings. Still, they are custom-made deep neural networks and sensitive to data noise, yielding limited performance in unseen real or fake facial images. In this paper, we aim to develop a simple yet effective deepfake detection method to improve the generalization abilities in unseen scenarios. To achieve this, we transfer the powerful classification abilities of general images of the ViT model to the downstream deepfake detection task by developing a parameter-efficient fine-tuning scheme with only a small number of trainable parameters and a simple structure.

\subsection{Parameter Efficient Fine-tuning}
Parameter-efficient fine-tuning (PEFT) methods~\cite{r50,r49,r54,r55} are attracting growing attention as they can efficiently adapt large-scale models to downstream tasks by fine-tuning only a small number of parameters.  
Motivated by the success of prompting in natural language processing (NLP), a recent work~\cite{r50} proposes Visual Prompt Tuning (VPT) that injects a small number of task-specific learnable parameters into the input sequence while keeping the Transformer backbone frozen. Another remarkable progress for PEFT is Adapter~\cite{Adapter2019}, which is integrated into transformer layers while maintaining its original parameters frozen. This adapter layer is composed of a down-projection linear layer, a nonlinear activation function, and an up-projection linear layer. DeepFake-Adapter~\cite{deepake-adapter2023} incorporates globally aware and locally aware adapters to detect forged clues. Although the DE-Adapter~\cite{de-adapter2024} proposes a novel parameter-efficient adaptation strategy and enhances detailed representations, it lacks cross-dataset generalization capability. Previous works\cite{r21,r22} have demonstrated that pretrained large-scale models essentially reside in a low intrinsic dimension space and can effectively learn even after being randomly projected onto a smaller subspace. As we know, LoRA~\cite{lora2022} designs the low-rank matrix approximation while freezing the original parameters of the large-scale language model. ViT+LoRA~\cite{vit_lora-2023} directly combines LoRA and ViT to detect manipulated facial images while lacking rational analysis and generalization ability on cross-manipulation. Thus, our method first learns more discriminative representations of real and synthetic facial pictures by capturing the continuity across patches in a real image and discontinuity in a forged image within a low-dimensional latent space. Moreover, we propose a class-invariant feature augmentation (CIFAug) function without adding additional trainable parameters to improve the generalization performance on unseen scenarios.

\subsection{Feature Modulation and Augmentation}
Data augmentation methods usually aim to regularize deep networks. Some normalization methods~\cite{2016layer,ioffe2015batch,wu2018group} usually first normalize feature representations and then apply learnable scale and shift transformations to achieve better performance in some tasks. 
The studies most related to our feature space redistribution are~\cite{devries2017dataset, tseng2020cross, lian2022scaling}. DeVries \& Taylor~\cite{devries2017dataset} apply a domain-agnostic approach to dataset augmentation in a learned feature space. To achieve this, they introduce noise $X$ by drawing from a Gaussian distribution with zero mean and per-element standard deviation calculated across all context vectors of the whole dataset as well as sets a hyper parameter used to scale the noise globally, and also apply interpolation by searching $K$ nearest neighbours and extrapolation operation to complete the data augmentation. The augmentation pipeline is complex. Tseng et al.~\cite{tseng2020cross} insert the feature-wise transformation layers to modulate intermediate feature activation $z$ in the feature encoder $E$ at multiple levels and propose a learning-to-learn algorithm to optimize the corresponding hyperparameters. Similarly, SSF~\cite{lian2022scaling} incorporates scaling and shifting parameters into every operation of an encoder $E$ to modulate the feature distributions. However, these two data augmentation methods both introduce a large number of trainable parameters, making the model architecture more complex. In particular, the former requires a carefully designed optimization algorithm for each affine parameter estimation. The key consideration of our feature space redistribution (FSR) is to improve the feature distributions with an excellent trade-off between performance and efficiency. To achieve this, we integrate an affine transformation to dynamically optimize the feature spaces produced by ViT once to improve the classification decision boundary between real and fake faces.

In addition, ISDA~\cite{2021regularizing} first estimates the covariance matrix of features for each class to capture intra-class variations. Then, it samples meaningful semantic transformation distributions from a multivariate normal distribution with zero mean and the estimated covariance and augments features with the distributions. Our CIFAug first computes the discriminative direction between real and fake classes and then computes the directions orthogonal to it to generate diversified real and forged features.
Unlike ISDA, our CIFAug does not consider which directions correspond to meaningful semantic transformations for the face and enforces estimating each class-specific covariance matrix, as some facial attributes are implicit and cannot be explicitly disentangled. Moreover, some forged faces may contain noticeable artifacts, such as significant skin tone discrepancies between manipulated regions and the original face, rendering strict semantic alignment impractical. Furthermore, our CIFAug computes the directions orthogonal to the classification direction rather than estimating the class-specific covariance matrices to determine the augmentation directions.

\begin{figure*}[t]
\centering
\includegraphics[width=1.0\linewidth]{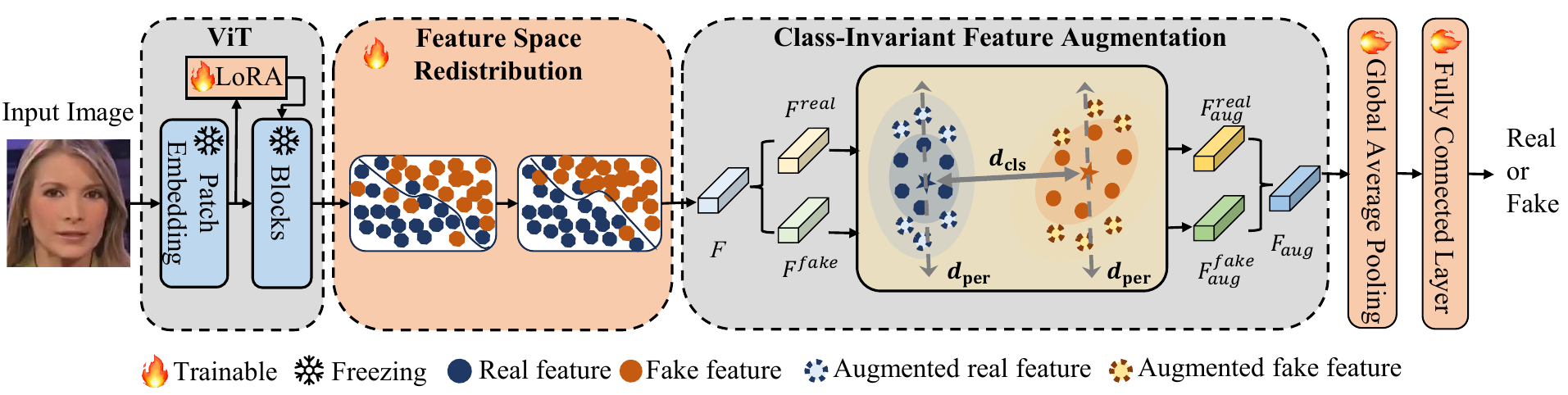}
\caption{The overall pipeline of the proposed generalizable deepfake detection framework (GenDF). The facial images (real or fake) first enter a ViT backbone for representation embedding in a low-dimensional space. Then, we optimize the distributions to learn more discriminative features. Next, these two kinds of features go through the class-invariant feature augmentation procedure to improve the generalization abilities of our method.}
\label{main}
\end{figure*}

%% file: section/3_Method.tex
\section{Methodology}
As shown in Fig.~\ref{main}, our proposed GenDF consists of three main schemes, including deepfake-specific representations learning (DSRL), feature space redistribution (FSR), and class-invariant feature augmentation (CIFAug) function. Following the same feature embedding procedure of ViT, our DSRL models an input image as a sequence of fixed-size patches and learns different relationships of real or fake patches by fine-tuning ViT in a low-dimensional space. Then, we introduce the feature space redistribution (FSR) modulation to separately optimize the distribution of real and fake feature spaces, enabling more discriminative representations. Furthermore, we propose a class-invariant feature augmentation (CIFAug) method that actively seeks diversified representations along class-invariant directions to improve the generalization ability in unseen fake scenarios.

\subsection{Deepfake-Specific Representations Learning} \label{sec:3.1}
 We assert that real facial images inherently exhibit continuity between patches, while forged facial images contain subtle artifacts, leading to discontinuity across patches. To capture these distinct relationships and accelerate convergence, we transfer ViT's prior knowledge from generic image classification to deepfake detection by fine-tuning it in a low-dimensional space. To achieve this, we integrate Low-Rank Adaptation (LoRA)\cite{lora2022} into each multi-head self-attention module of ViT, enabling the model to learn deepfake-specific representations between patches within low-rank subspaces for real and fake facial images. Notably, this approach introduces only a small number of trainable parameters while keeping the pretrained ViT parameters frozen. 

Specifically, we employ separate low-rank decompositions, $B_{1}A_{1}$ and $ B_{2}A_{2}$, to learn $Q$ and $V$ matrices, allowing the ViT image encoder to learn different patch relationships in real and forged faces. Furthermore, we freeze the pre-trained weights of the $K$ matrix to retain the foundation knowledge embedded in ViT. However, to effectively learn the internal relationships within facial images (real or fake), it is essential to fine-tune the $V$ matrix to acquire deepfake-specific knowledge.

Formally, the self-attention mechanism of ViT comprises three weight matrices, namely query $Q$, key $K$, and value $V$, formulated as: 
\begin{equation}
     Q=W_{q}x,  K=W_{k}x,  V=W_{v}x,
\end{equation}
where $x \in \mathbb{R}^{H \times W \times D}$ represents feature map from the preceding layer, and $(W_{q}, W_{k}, W_{v}) \in \mathbb{R}^{D \times D}$ are the pretrained weights. To achieve low-rank adaptation, we introduce trainable parameters $\Delta W_{q}$ and $\Delta W_{v}$ into $Q$ and $V$, respectively, within each self-attention layer of ViT's encoder, while keeping the original pretrained weights $W_{q}$, $W_{k}$, and $W_{v}$ frozen. 
With the integration of low-rank adaptation, the query Q and value V are updated as:
\begin{equation}
\begin{aligned}
    Q &= W_{q}x + \Delta W_{q}x=W_{q}x + B_{1}A_{1}x\\
    V &= W_{v}x + \Delta W_{v}x=W_{v}x+B_{2}A_{2}x,
\end{aligned}
\end{equation}
where the low-rank decompositions of $\Delta W_{x}$ are given by $\Delta W_{x}=B_{x}A_{x}$, specifically $\Delta W_{q}=B_{1}A_{1}$ and $\Delta W_{v}=B_{2}A_{2}$. Here, $(B_{1},B_{2}) \in \mathbb{R}^{D \times r}$, $(A_{1},A_{2}) \in \mathbb{R}^{r \times D}$, with rank $r$ is given sufficiently low. During training, only the low-rank matrices (i.e., $B_{1}$, $A_{1}$, $B_{2}$ and $A_{2}$) are optimized. In practice, we initialize the product $BA$ to zero at the beginning of the fine-tuning process by setting $B=0$ and sampling $A$ from a normal distribution, $A\sim {N(0,\sigma^ {2}})$.

\subsection{Feature Space Redistribution}
While the pretrained ViT exhibits powerful recognition ability on generic camera-captured images, it lacks basic knowledge on the domain of real and fake human faces. However, fine-tuning ViT with the DSRL struggles to match the domain gap completely. To further address the distribution mismatch between generic images and humans' real and fake facial images, we propose Feature Space Redistribution (FSR) to optimize the distributions of real and forged features separately and increase the inter-class distance, facilitating better discrimination. In contrast to SSF~\cite{lian2022scaling} and the work~\cite{tseng2020cross}, which require feature-wise transformation at every operation or encoder layer along with a carefully crafted complex hyperparameter optimization, the proposed FSR applies a scaling factor and noise only once to the output of the fine-tuned ViT and optimizes the parameters directly through gradient descent without additional computation. 
\\

Concretely, given a feature (real or fake) from the fine-tuned ViT, we first multiply it using a multiplication factor sampled from a Gaussian distribution to reduce or amplify each dimension individually and then introduce a noise parameter drawn from a Gaussian distribution to mimic uncertainty. Physically, the proposed FSR adaptively optimizes the distribution scopes of real and fake features separately and increases the inter-class distance between the two classes, facilitating better discrimination.

We denote the feature of the ViT encoder by $x \in \mathbb{R}^{N \times d}$. The above feature modulation process can be calculated as follows: 
\begin{equation}\label{eq: FSR}
	F = \theta \odot x + \epsilon,
\end{equation}
where $\theta \in \mathbb{R}^{d}$ is the multiplication factor, $\epsilon \in \mathbb{R}^{d}$ is the noise, and $\odot$ denotes the element-wise multiplication on feature dimension. Here, $\theta\sim {N(1,\sigma ^ {2}})$ and $\epsilon\sim {N(0,\sigma_{\epsilon} ^ {2}})$. The modulated feature $F \in \mathbb{R}^{N \times d}$ serves as the input of the next module of our method.

\subsection{Class-Invariant Feature Augmentation}
In the real world, real facial images and videos are influenced by contrast, lighting, ethnicity, facial angles, etc., resulting in diverse distributions. Moreover, the continuous advancement of deepfake techniques constantly leads to the emergence of new types of forgeries. These factors further enhance the difficulty of distinguishing real and fake faces in unseen domains. To this end, we propose a Class-Invariant Feature Augmentation (CIFAug) function to enhance generalization by encouraging diverse representations along class-invariant directions. Specifically, CIFAug expands the feature space scopes of both real and fake faces along directions orthogonal to the classification direction, enabling the model to capture a broader range of variations. These augmented features effectively improve the model's generalization ability on unseen real or forged patterns.

Specifically, given the feature map $F \in \mathbb{R}^{N \times 197 \times C}$ produced by the proposed FSR, we first pool $F$ into feature vectors $f \in \mathbb{R}^{N \times C}$, followed by splitting $f$ into real feature $f ^ {real} \in \mathbb{R}^{\frac{N}{2} \times C}$ and fake feature $f ^ {fake} \in \mathbb{R}^{\frac{N}{2} \times C}$ according to corresponding binary labels. Here, $197$ represents $14 \times 14$ patches and $1$ class token, $N$ is the batch size, and $C$ is the embedding dimension. Subsequently, we estimate the category change direction $\widetilde{d}_{cls}$ by computing the vector from the centroid of real faces to that of fake faces. $\widetilde{d}_{cls}$ is then normalized
to obtain the unit vector of the discriminating category ${d}_{cls}$:
\begin{equation}
\begin{aligned}
    \widetilde{d}_{cls} &= \Bar{f} ^ {real} - \Bar{f} ^ {fake}\\
    d_{cls} &= \frac{\widetilde{d}_{cls}}{\| \widetilde{d}_{cls} \|_2},
\end{aligned}
\end{equation}
where $\Bar{f}$ denotes the average vector of feature vectors $f$ and $\| \widetilde{d}_{cls} \|_2$ is the $l_2$ norm of  $\widetilde{d}_{cls}$. To diversify the class-invariant information of real and fake faces, we apply the Gram-Schmidt Orthogonalization to compute the direction ${d}_{per}$ that is orthogonal to the discriminative direction ${d}_{cls}$. Specifically, we first initialize a random direction $d_{rand}\sim {N(0,1})$ and then compute:
\begin{equation}
\begin{aligned}
    \widetilde{d}_{per} &= d_{rand} - \frac{\langle d_{rand},d_{cls} \rangle}{\langle d_{cls},d_{cls} \rangle} \cdot {d_{cls}}\\
    {d}_{per} &= \frac{\widetilde{d}_{per}}{\| \widetilde{d}_{per} \|_2},
\end{aligned}
\end{equation}
where ${\langle \cdot,\cdot  \rangle}$ denotes the inner-product operation, which ensures that $d_{per}$ stems from the direction independent of class-specific changes.
Hence, we adopt $d_{per}$ to augment the original features to obtain diversified fake and real features:
\begin{equation}
\begin{aligned}
     F_{syn} ^ {real} &= F ^ {real} + \beta \cdot {\| f ^ {real} \|_2} \cdot{d}_{per}\\
     F_{syn} ^ {fake} &= F ^ {fake} + \beta \cdot {\| f ^ {fake} \|_2} \cdot{d}_{per},
\end{aligned}     
\end{equation}
where $\beta \sim U(-\sigma, \sigma)$ and $\sigma$ is a hyper-parameter to control the scale of the perturbation ($\sigma$ = 0.02 by default). Synthesized features $F_{syn}^{real}$  and $F_{syn}^{fake}$ are concatenated into $F_{aug}$, and then the augmented feature $F_{aug}$ is sent to the classification head of the GenDF. Note that the feature augmentation process is activated with a probability of 0.5 in the training stage and is deactivated at inference time.

\subsection{Objective Function}
\subsubsection{Cross-Entropy Loss}
We employ the cross-entropy loss $\mathcal{L}_{ce}$ to guide the model in distinguishing between the two classes. $\mathcal{L}_{ce}$ is formulated as:
\begin{equation}
    \mathcal{L}_{ce} = - \frac{1}{N}\Big(\sum\limits_{i=1}^{N} y_{i}log(\hat{y_{i}}) + \hat{y_{i}}log(y_{i})\Big),
\end{equation}
where $N$ is the number of input images within a mini-batch of training samples, $y_{i}$ and $\hat{y_{i}}$ are the ground-truth label and the corresponding prediction, respectively.

\vspace{0.05in}
\subsubsection{Triplet Loss}
To learn more discriminative representations, we introduce a weighted triplet loss $\mathcal{L}_{tri}$ to aggregate real faces while pushing away various fake attacks from real samples. Consequently, the learned representations of real facial images are more likely to help recognize various forged samples as outliers. Given the pooled feature vector $f$, the proposed weighted triplet loss is formulated as:
\begin{equation}
\begin{aligned}
    \mathcal{L}_{tri} &= \delta \Big(\sum_{f_p} w_{p}d(f_a,f_p)-\sum_{f_n} w_{n}d(f_a,f_n)\Big)\\
    w_p &= \text{Softmax}\left(d(f_a,f_p)\right)\\
    w_n &= \text{Softmax}\left(-d(f_a,f_n)\right),
\end{aligned} 
\end{equation}
where $f_a$, $f_p$, and $f_n$ denote the features of anchor, positive, and negative samples, respectively. Here, $\delta$ is the softplus function defined as $\delta(\cdot) = ln(1+exp(\cdot))$. $d(\cdot,\cdot)$ is any distance metric and is chosen as the Euclidean distance in our method. Here, $w_p$ and $w_n$ act as weighting coefficients that dynamically focus on harder samples according to instance-wise confidence, thereby facilitating the learning of more discriminative classification boundaries. The softplus function $\ Delta (\cdot)$ replaces the ReLU function used in the hard margin triplet loss\cite{Schroff_2015}, eliminating the need for margin hyperparameter tuning. In essence, $\mathcal{L}_{tri}$ promotes a compact feature space for real faces while pushing fake faces away, thereby establishing a clear decision boundary between real and fake samples, ultimately enhancing generalization.

\begin{table*}[t]
     \renewcommand{\arraystretch}{1.2} 
  \caption{Cross-Domains Evaluations by Training on FF++(HQ)~\cite{FF++xception-iccv2019} and Evaluating on Other Datasets in Terms of Frame-Level AUC ($\%$) and EER Metrics. The Best and Second Results Are Highlighted in Bold and \underline{Underline} Font. * Denotes That We Reproduce the Official Code.
  }
  \label{tab:1}
  \centering
  \setlength{\tabcolsep}{1.0mm}{
  \begin{tabular}{llccccccccc}
    \toprule
    \multirow{2}{*}{Categories} & \multirow{2}{*}{Methods} &\multirow{2}{*}{\makecell[c]{\# Trainable \\ Parameters}}  &\multicolumn{1}{c}{FF++} & \multicolumn{2}{c}{Celeb-DF} 
    &\multicolumn{2}{c}{DFDC} &\multicolumn{2}{c}{DFD} \\ 
    \cmidrule{4-10}  & & &AUC (\%) $\uparrow$ &AUC (\%) $\uparrow$ &EER $\downarrow$    &AUC (\%) $\uparrow$  &EER $\downarrow$   &AUC (\%) $\uparrow$  &EER $\downarrow$ \\ 
    \midrule
     \multirow{6}{*}{Custom-made} 
     &$F^{3}$-Net\cite{F3-net} &20.75M &98.10 & 61.51 & 0.420 &64.60 &0.398  &79.75  &{-}  \\
      &RFM~\cite{RFM2021}  &{-} &98.79  &65.63 &0.385  &66.01 &0.391 &{-} &{-} \\
     &MultiAtt\cite{zhao2021multi} &417.52M  &99.29 & 67.02 & 0.379 &68.01 &0.372    &{-}   &{-} \\
    &RECCE*\cite{recce-cvpr2022} &24.69M &97.79 &68.71   &0.357 &69.06  &0.361  &81.19 &{-} \\
    & UIA-ViT*\cite{uia-vit-eccv2022} &81.83M &98.26 &\underline{80.47}  &{-} &71.84 &{-}  &81.96 &{-}\\
    &UCF\cite{ucf-iccv2023} &46.84M &97.05 &75.27   &{-} &71.91  &{-}  &80.74 &{-} \\
    \midrule
    \multirow{4}{*}{Pre-trained}    &Xception\cite{xception} &20.81M  &96.37 &61.80 & 0.417 &63.61 &0.406  &81.63   &{-} \\
     &EfficientNet-B4\cite{Efficientnet-icml2019} &19M &99.18 &74.87 &{-} &69.55 &{-}  &81.48   &{-} \\
    & ViT*\cite{vit-iclr2021} &1.5K &78.45 &66.40 &0.388 &59.64 &0.433  &69.03 &0.356 \\
     \midrule
    \multirow{5}{*}{Fine-tuned}  & ViT+LoRA\cite{vit_lora-2023} &0.27M &{-} &79.67  & {-}   &71.74     &{-}     &\underline{83.42}  &{-} \\
    &Deepfake+Adapter\cite{deepake-adapter2023} &16.92M &98.73 & 71.74 &0.340    &\underline{72.66}    &\underline{0.327}   &{-}   &{-} \\
    &DE-Adapter\cite{de-adapter2024}  &0.38M  &\textbf{99.31} &67.12   &{-}   &68.84  &{-} &{-} &{-} \\
    &GenDF (Ours) &0.28M  &\textbf{99.31} &\textbf{82.00}  &\textbf{0.258} &\textbf{73.24}  &\textbf{0.304} &\textbf{86.53} &\textbf{0.221} \\
  \bottomrule
  \end{tabular}}
 \end{table*}

\vspace{0.05in}
\subsubsection{Feature Augmentation Loss}

we further employ a feature augmentation loss $\mathcal{L}_{aug}$, leveraging the same cross-entropy loss as $\mathcal{L}_{ce}$ to optimize the CIFAug procedure. However, the difference between $\mathcal{L}_{ce}$  and $\mathcal{L}_{aug}$ lies in the fact that the predicted value $\hat{y_{i}}$ in $\mathcal{L}_{ce}$ is derived from the classification head using original features $x$ without augmentation, while $\hat{y_{i}}$ in $\mathcal{L}_{aug}$ is obtained from the classification head based on the augmented feature $F_{aug}$. Notably, the feature augmentation loss is set to zero with a 50\% probability due to our random decision-making process, which involves a 50\% possibility of whether to incorporate the CIFAug strategy.

\vspace{0.05in}
\subsubsection{Overall Loss}
The overall objective function $\mathcal{L}$ of the proposed method is a weighted sum of the cross-entropy loss $\mathcal{L}_{ce}$, the triplet loss $\mathcal{L}_{tri}$ and the feature augmentation loss $\mathcal{L}_{aug}$:
\begin{equation}
    \setlength{\abovedisplayskip}{3pt}
    \mathcal{L} = \mathcal{L}_{ce} + \gamma_1 \mathcal{L}_{tri} + \gamma_2 \mathcal{L}_{aug},
    \setlength{\belowdisplayskip}{3pt}
    \label{eq:total_loss}
\end{equation}
where $\gamma_1$ and $\gamma_2$ are weight parameters for balancing different losses. \\

%% file: section/4_Experiment.tex
\section{Experiment}
In this section, we first introduce the experiment setting, including datasets, competing deepfake detection methods, and implementation details. Then, we compare our method with existing deepfake detectors in cross-domain, cross-manipulation, and within-domain settings to thoroughly investigate the quantitative performance of our GenDF. Furthermore, we conduct the qualitative analysis by visualizing the feature distribution and patch-discontinuity patterns within fake faces and patch-continuity within real faces captured by our method, providing a more intuitive demonstration of GenDF's effectiveness. Then, we conduct diverse ablation experiments to validate the efficacy of individual components and parameters within the proposed GenDF. We also evaluate the robustness of our method against various disturbances in image quality. 

\subsection{Experimental Setting}
\label{sec: Experimental Setup}

\subsubsection{Datasets}
Following existing deepfake detection works~\cite{FF++xception-iccv2019, zhao2021multi, F3-net}, we evaluate performance of the proposed GenDF on four large-scale benchmarks: FaceForensics++ (FF++)~\cite{FF++xception-iccv2019}, Celeb-DF~\cite{r29}, Deepfake Detection Challenge (DFDC)~\cite{r28}, and DeepfakeDetection (DFD)~\cite{dfd}. 

Specifically, FF++ consists of 1,000 original (real) videos from YouTube, along with four types of manipulation techniques, including Deepfakes (DF), Face2Face (F2F), FaceSwap (FS), and NeuralTextures (NT). The dataset is available in raw, high-quality (HQ), and low-quality (LQ) versions. Since facial images typically exhibit limited visual quality, we follow~\cite{recce-cvpr2022, zhao2021multi,ucf-iccv2023} and adopt the HQ version with a 23\% compression ratio in our experiments instead of the raw version. This choice strikes a balance between model performance and robustness. The Celeb-DF dataset contains 590 real videos captured in various scenarios and 5,639 fake videos. The forged sequences are manipulated using an improved DeepFake technique. DFDC is a challenging, large-scale dataset with 128,154 video clips containing 23,654 real videos and 104,500 fake videos. The real faces of DFD are collected from self-recorded movies shot in natural environments, indoors and outdoors, including a variety of real-world light conditions. The fake videos in DFD are created using several Deepfake, GAN-based, and non-learned methods. We separately divide these four datasets into training, validation, and test sets for fair and accurate performance comparisons of state-of-the-art techniques and our approach.

\vspace{0.05in}
\subsubsection{Implementation Details}
 We fine-tune the ViT model using the proposed GenDF on the training set, select the optimal model parameters based on the validation set, and evaluate its performance on the test set. The datasets FF++, Celeb-DF, DFDC, and DFD are split into training, validation, and test sets following an 8:1:1 ratio, respectively. We apply the ViT-B/16 as our backbone, which consists of 12 transformer blocks with the input size of $224 \times 224$ and the patch size  $16 \times 16$. The batch size is set to 96. We adopt the AdamW~\cite{kingma2014adam} optimizer with a learning rate of 3e-4. In Eqn. \eqref{eq:total_loss}, $\gamma_1$, $\gamma_2$ are both set to 0.2. The rank $r$ of LoRA is set to 8, and we scale  $\Delta W_{q}$ and $\Delta W_{v}$ by a factor of 4. To assess the effectiveness of our method, we utilize common metrics in deepfake detection, including Accuracy score (Acc), Area Under the Receiver Operating Characteristic Curve (AUC), and Equal Error Rate (EER). We report the video-level results on FF++ and the frame-level results on other datasets. Our method is trained using 4 Nvidia RTX4090 GPUs.

\vspace{0.05in}
\subsubsection{Competing Methods}
\label{sec:4.1.3}
We classify existing deepfake detection methods into three categories: (1) custom-made methods incorporating various prior knowledge through complex architectures, (2) pre-trained models on generic images, and (3) fine-tuned large-scale vision models with a small number of trainable parameters. Custom-designed models include $F^{3}$-Net~\cite{F3-net}, RFM~\cite{RFM2021}, MultiAtt~\cite{zhao2021multi}, RECCE~\cite{recce-cvpr2022}, UIA-ViT~\cite{uia-vit-eccv2022}, and UCF~\cite{ucf-iccv2023}. These models achieve promising performance but involve complex neural network architectures to incorporate prior knowledge. However, they suffer significant performance degradation when encountering unseen forged patterns. Notably, UIA-ViT fully fine-tunes ViT and incorporates the prior knowledge of real and fake data following a Gaussian distribution, resulting in a complex feature learning process. Additionally, Xception~\cite{xception} and EfficientNet-B4~\cite{Efficientnet-icml2019}, which are pre-trained on ImageNet~\cite{deng2009imagenet}, serve as the backbones for numerous deepfake detection methods. As our baseline model, we utilize a pre-trained ViT~\cite{vit-iclr2021} with a defined classification head comprising an average pooling layer and a fully connected layer. We merely train the classification head while keeping all pre-trained parameters frozen. Finetuning-based methods optimize some parameters while freezing most pre-trained parameters via parameter-efficient fine-tuning (PEFT) mechanisms. Recent methods in this category include ViT+LoRA~\cite{vit_lora-2023}, Deepfake+Adapter~\cite{deepake-adapter2023}, and DE-Adapter~\cite{de-adapter2024}, which fine-tune large-scale vision models for deepfake detection. To ensure fair comparisons, we reproduce the results of RECCE, UIA-ViT, and ViT using their official codes and parameters. The result of UCF is sourced from Deepfakebench~\cite{Deepfakebench-2023}. Additionally, we report the results of EfficientNet-B4, $F^{3}$-Net, MultiAtt, ViT+LoRA, Deepfake+Adapter, and DE-Adapter as presented in their papers.

\subsection{Experimental Results}
\subsubsection{Cross-Domain Evaluations}
To assess the generalization capability of our method on unseen forged facial images, we adhere to the common paradigm of training on the FF++(HQ)~\cite{FF++xception-iccv2019} dataset and testing on other datasets, including Celeb-DF, DFDC, and DFD. 

As demonstrated in Table~\ref{tab:1}, our method achieves state-of-the-art generalization performance in cross-domain settings. Specifically, the GenDF remarkably improves AUC scores on DFD, Celeb-DF, and DFDC datasets by 1.53\%, 0.58\%, and 3.11\%, respectively. Notably, our GenDF fine-tunes only 0.34\% of the parameters of the fully trained UIA-ViT~\cite{uia-vit-eccv2022} yet consistently outperforms all fine-tuning, pretrained, and custom-designed methods. Additionally, while the ViT-B/16 backbone has lower performance compared to CNN-based pre-trained models such as Xception~\cite{xception} and EfficientNet-B4~\cite{Efficientnet-icml2019}, incorporating GenDF enables ViT to achieve state-of-the-art performance. Moreover, our framework significantly surpasses ViT+LoRA~\cite{vit_lora-2023} across all unseen manipulated scenarios, indicating that directly applying LoRA for ViT fine-tuning without rational analysis is insufficient to capture deepfake-specific representations. In contrast, GenDF effectively distinguishes real from fake facial images by fine-tuning ViT while preserving prior knowledge of generic images and augmenting features. These findings demonstrate that GenDF efficiently captures the distinction between real and fake facial images only with a small number of trainable parameters, highlighting its simplicity and strong generalization capability in cross-domain settings.  

\vspace{0.05in}
\begin{table}[t]
 \setlength{\tabcolsep}{4pt}
 \renewcommand{\arraystretch}{1.2}
  \caption{Evaluation Performance on Cross-Manipulation Settings in Terms of AUC on FF++~\cite{FF++xception-iccv2019} Dataset. Cells in gray denote the intra-manipulation results. The "Avg." represents the Mean score. The Best Results Are Highlighted in Bold Font.}
  \label{tab:2}
  \centering
  \begin{tabular}{lcccccc}
    \toprule
    \multirow{2}{*}{Methods}   &\multirow{2}{*}{Training } &\multicolumn{5}{c}{Testing} \\
    \cmidrule{3-7} &  &DF &F2F &FS &NT &Avg. \\
    \midrule
    Xception~\cite{xception} &\multirow{7}{*}{\quad DF \quad} &\cellcolor{gray!30}0.994 &0.723 &0.703 &0.757 &0.794 \\
    $F^{3}$-Net~\cite{F3-net} & &\cellcolor{gray!30}0.976 &0.637 &\textbf{0.734} &0.695 &0.761 \\
    EfficientB4~\cite{Efficientnet-icml2019} & &\cellcolor{gray!30}0.995 &\textbf{0.728} &0.706 &\textbf{0.771} &\textbf{0.800} \\
    SRM~\cite{srm2021} & &\cellcolor{gray!30}0.995 &0.724 &0.684 &0.762 &0.791 \\
    UIA-ViT~\cite{uia-vit-eccv2022} & &\cellcolor{gray!30}0.994 &0.636 &0.579 &0.634 &0.711 \\
    GenDF (Ours) & &\cellcolor{gray!30}\textbf{0.999} &0.640 &0.661 &0.690 &0.748 \\ 
    \hline
    
    Xception~\cite{xception} &\multirow{7}{*}{\quad F2F \quad} &0.674 &\cellcolor{gray!30}0.984 &0.633 &0.679 &0.743 \\
    $F^{3}$-Net~\cite{F3-net} & &0.601 &\cellcolor{gray!30}0.962 &0.583 &0.656 &0.701 \\
    EfficientB4~\cite{Efficientnet-icml2019} & &0.661 &\cellcolor{gray!30}0.982 &0.629 &0.682 &0.739 \\
    SRM~\cite{srm2021} & &0.711 &\cellcolor{gray!30}0.985 &0.622 &0.680 &0.750 \\
    UIA-ViT~\cite{uia-vit-eccv2022} & &0.774 &\cellcolor{gray!30}0.990 &0.615 &0.582 &0.740 \\
    GenDF (Ours)  & &\textbf{0.807} &\cellcolor{gray!30}\textbf{0.992} &0.613 &0.627 &\textbf{0.760} \\
    \hline
    
    Xception~\cite{xception} &\multirow{7}{*}{\quad FS \quad} &0.552 &0.609 &\cellcolor{gray!30}0.993 &0.551 &0.676 \\
    $F^{3}$-Net~\cite{F3-net} & &0.643 &0.582 &\cellcolor{gray!30}0.975 &0.524 &0.681 \\
    EfficientB4~\cite{Efficientnet-icml2019}  & &0.555 &0.596 &\cellcolor{gray!30}0.992 &0.553 &0.674 \\
    SRM~\cite{srm2021} & &0.550 &0.607 &\cellcolor{gray!30}0.993 &0.550 &0.675 \\
    UIA-ViT~\cite{uia-vit-eccv2022} & &\textbf{0.885} &0.634 &\cellcolor{gray!30}0.995 &0.477 &0.748 \\
    GenDF (Ours) & &0.827 &\textbf{0.656} &\cellcolor{gray!30}\textbf{0.999} &\textbf{0.565} &\textbf{0.762} \\ \hline
     
     Xception~\cite{xception} &\multirow{7}{*}{\quad NT \quad}  &0.635 &0.585 &0.535 &\cellcolor{gray!30} 0.955 &0.678 \\
    $F^{3}$-Net~\cite{F3-net} & &0.580 &0.562 &0.523 &\cellcolor{gray!30}0.846 &0.628 \\
    EfficientB4~\cite{Efficientnet-icml2019} & &0.649 &0.584 &0.525 &\cellcolor{gray!30}0.956 &0.679 \\
    SRM~\cite{srm2021} & &0.649 &0.578 &0.530 &\cellcolor{gray!30}0.955 &0.678 \\
    UIA-ViT~\cite{uia-vit-eccv2022} & &0.729 &0.678 &0.461 &\cellcolor{gray!30}\textbf{0.963} &0.708 \\
    GenDF (Ours) & &\textbf{0.747} &\textbf{0.685} &\textbf{0.542} &\cellcolor{gray!30}0.957 &\textbf{0.733} \\ 
  \bottomrule
  \end{tabular}
\end{table}

 \begin{table*}[t]
 \renewcommand{\arraystretch}{1.2}  
 \setlength{\tabcolsep}{1.6mm}{
  \caption{The Comparison of Intra-Domain Performance about FF++~\cite{FF++xception-iccv2019}, Celeb-DF~\cite{r29}, DFDC~\cite{r28} and DFD~\cite{dfd}. We Report Video-Level Acc ($\%$) and AUC ($\%$) on FF++ and Frame-Level Acc ($\%$) and AUC ($\%$) on Other Datasets. The Best and Second Results Are Highlighted in Bold and \underline{Underline} Font, Respectively.}
  \label{tab:3}
  \centering
  \begin{tabular}{@{}clcccccccccccc@{}}
    \toprule
      \multirow{2}{*}{Categories} &\multirow{2}{*}{Methods}  &\multicolumn{2}{c}{FF++} 
     &\multicolumn{2}{c}{Celeb-DF} &\multicolumn{2}{c}{DFDC}  &\multicolumn{2}{c}{DFD}   \\\cmidrule{3-10} &  &Acc (\%) $\uparrow$ 
      & AUC (\%) $\uparrow$    & Acc (\%) $\uparrow$         & AUC (\%) $\uparrow$ & Acc (\%) $\uparrow$         & AUC (\%) $\uparrow$ & Acc (\%) $\uparrow$         & AUC (\%) $\uparrow$\\
    \midrule
    \multirow{6}{*}{\quad Custom-made \quad} 
     &$F^{3}$-Net~\cite{F3-net}   & 97.52 & 98.10    &95.95 &98.93 &76.17 &88.39 &{-} &{-}\\
     &MultiAtt~\cite{zhao2021multi}   &97.60 &\underline{99.29}  &97.92 &\textbf{99.94} &76.81 &90.32 &{-} &{-}\\
     &RFM~\cite{RFM2021}   &95.69 &98.79  &97.96 &\textbf{99.94} &80.83 &89.75 &{-} &{-} \\
     &Add-Net~\cite{r30Add-Net} &96.78 &97.94 &96.93 &99.55 &78.71 &89.85 &{-} &\underline{97.51} \\
     &RECCE*~\cite{recce-cvpr2022} &96.98   &99.29  &\underline{98.59} &\textbf{99.94} &81.20 &\underline{90.58} &{-} &{-}\\
     & UIA-ViT*~\cite{uia-vit-eccv2022}      &95.71 & 98.25  &{98.52} &{98.84} &{80.23} &{89.98} &{-} &94.68\\
     \midrule
    \multirow{2}{*}{Pre-trained} &Xception~\cite{xception}   & 95.73 &96.37    &97.90 &99.73 &79.35 &89.50 &{-} &{-}\\
    & ViT*\cite{vit-iclr2021}    &73.57    &78.45    &84.17 &87.19  &64.09 &68.74 &79.06 &84.01\\
   
    \midrule
    \multirow{2}{*}{Fine-tuned}
    &DE-Adapter\cite{de-adapter2024}    &\textbf{98.01}  &\textbf{99.31}    &\textbf{98.93} &\underline{99.91} &\underline{81.59} &90.37 &{-} &{-}\\
     &GenDF (Ours)   &\underline{97.72}  &\textbf{99.31}  &\underline{98.82} &\textbf{99.94} &\textbf{84.05} &\textbf{91.67} &\textbf{94.14} &\textbf{98.26}\\
    
  \bottomrule
  \end{tabular}}
\end{table*}

\begin{table}[t]
     \setlength{\tabcolsep}{6pt}   
     \renewcommand{\arraystretch}{1.2}  
  \caption{The evaluation results of the efficiency and effectiveness trade-off of our method and baselines. The Avg. represents the average test results of Celeb-DF, DFDC, and DFD on cross-domain settings trained on FF++(HQ).}
  \label{tab:10}
  \centering
  \setlength{\tabcolsep}{1.1mm}{
  \begin{tabular}{lccccc}
    \toprule
    Methods &\makecell[c]{\#Params} &GFLOPs &\makecell[c]{Inference \\Time~(ms)} &\makecell[c]{Avg. \\ AUC~(\%)}\\\midrule
    Face-x-ray~\cite{r41-Face-x-ray} &77.47M &42.58 &35.62 &76.60 \\
    Xception~\cite{xception} &20.81M &16.84 &5.25 &69.01\\
    RECCE\cite{recce-cvpr2022} &24.69M &13.18 &99.37 &72.99\\
    UIA-ViT\cite{uia-vit-eccv2022} &81.83M &17.58 &8.69 &78.09\\
    ViT+Lora\cite{vit_lora-2023} &0.27M &17.58 &12.12 &78.28\\
    GenDF (Ours) &0.28M &17.58 &9.21 &\textbf{80.59}\\
  \bottomrule
  \end{tabular}}
 \end{table}

\subsubsection{Cross-Manipulation Evaluations}
We further conduct additional experiments on four different manipulation types in the FF++(HQ) dataset to evaluate the generalization ability of GenDF in cross-manipulation settings. We compare Ours+ViT with state-of-the-art models by training on one forgery type and testing on each of the four types. The results for Xception~\cite{xception}, EfficientB4~\cite{Efficientnet-icml2019}, $F^{3}$-Net~\cite{F3-net}, SRM~\cite{srm2021} are obtained from the DeepfakeBench~\cite{Deepfakebench-2023} benchmark, while UIA-ViT~\cite{uia-vit-eccv2022} is reproduced by using its official code. \\

As shown in Table~\ref{tab:2}, ours+ViT achieves the highest average generalization performance when trained on F2F, FS, and NT, respectively, and tested across all four forgery types. While EfficientB4~\cite{Efficientnet-icml2019} achieved an average AUC score of 0.800 when trained on DF types, its performance significantly dropped to 0.674 and 0.679 when trained on FS and NT, respectively. In contrast, our method maintains a more consistent result, achieving an average AUC score of 0.751 across all four forgery techniques, with a maximum score of 0.762 and a minimum score of 0.733. These results indicate that GenDF exhibits minimal variance across different forgery techniques, making it more robust. Furthermore, our approach outperforms UIA-ViT by 2.4\% in average AUC across the four manipulation types. In summary, our framework surpasses pretrained CNN-based models and the fully fine-tuned UIA-ViT, demonstrating superior generalization to unseen forgery patterns.

\subsubsection{Within-Domain Evaluations}
For deepfake detection, we focus on the generalization performance of unseen manipulated patterns and the robustness of detection models across diverse forgery types rather than focusing solely on the traditional FF++ dataset or the relatively small Celeb-DF dataset. Given the varying forgery distributions across datasets, we comprehensively evaluate our method on FF++, Celeb-DF, DFDC, and DFD, comparing its performance with state-of-the-art deepfake detection methods. The quantitative results are presented in Table~\ref{tab:3}. Our method achieves state-of-the-art performance across different domains containing real and forged facial images while fine-tuning only a small number of parameters. Notably, our method improves the AUC on the DFD dataset by 3.58\% over UIA-ViT and enhances the Acc on the DFDC dataset by 2.46\% compared to DE-Adapter. 
These results highlight that our simple fine-tuning framework efficiently transfers the classification abilities of pre-trained ViT on generic images to deepfake detection while using significantly fewer trainable parameters. The detailed comparison between the Adapter and our DSRL is shown in the section \ref{sec:ablation}. Additionally, our method achieves comparable performance on FF++ and Celeb-DF, where AUC scores of state-of-the-art methods are already close to 100\%. However, our framework fine-tunes only 0.28M parameters, which is approximately 100 times fewer than MultiAtt.

\subsubsection{Evaluation Results of Efficiency/Effectiveness Trade-off}
To evaluate the efficiency of our method, we further compare the number of trainable parameters, GFLOPs, single-frame inference time, and average generalization performance in cross-domain settings with some representative baselines, as summarized in Table~\ref{tab:10}. All experiments are conducted on an Nvidia RTX 4090 GPU. Quantitative comparison results in~\cref{tab:1,tab:2,tab:3} and Table~\ref{tab:7} have demonstrated that the generalization ability and robustness of our method outperforms the baselines. Although the inference time of our method (9.21 ms) is slightly higher than that of Xception (5.25 ms) and UIA-ViT (8.69 ms). However, it achieves significant 11.58\% and 2.50\% improvements in average generalization performance, respectively, while requiring only 1.35\% and 0.34\% of their trainable parameters. These results demonstrate that GenDF strikes a better balance between effectiveness and efficiency.

\begin{table}[tb]
 \setlength{\tabcolsep}{5pt} 
 \renewcommand{\arraystretch}{1.2} 
  \caption{Ablation Study on the Effect of Different Components on FF++~\cite{FF++xception-iccv2019} Dataset.  The Best Results in Terms of Acc ($\%$) and AUC ($\%$) Are Highlighted in Bold Font.
  }
  \label{tab:4}
  \centering
  \begin{tabular}{cccccccc}
    \toprule
    &ID   &ViT    &DSRL  &FSR  &CIFAug  & Acc (\%)  & AUC (\%) \\
    \midrule
    &(a)   &\ding{52} &  & &  &73.57 &78.45\\
    &(b)   &\ding{52}  &  &\ding{52} & \ding{52} &80.14 &81.57\\
    &(c)   &\ding{52}  & \ding{52} & & \ding{52} &93.00 &97.02\\
    &(d)  &\ding{52}  & \ding{52} &\ding{52} &  &94.43 &97.39\\
    &GenDF (Ours)  & \ding{52} &\ding{52}  & \ding{52} & \ding{52} &\textbf{97.72}  &\textbf{99.31}\\
  \bottomrule
  \end{tabular}
\end{table}
 
\subsection{Ablation Studies}\label{sec:ablation}
\subsubsection{The Effects of Individual Components}
In this section, we perform ablation experiments to evaluate the effectiveness of each key component in our method on the FF++(HQ) dataset. Specifically, we assess the following variants: (a) the baseline ViT model, which trains only the classification head comprising a global average pooling layer and a fully connected layer, (b) our proposed GenDF without deepfake-specific representations learning (DSRL) scheme, (c) GenDF without feature space redistribution (FSR), (d) GenDF without CIFAug, and (e) Ours+ViT model. 

As shown in Table~\ref{tab:4}, the variant ViT with FSR and CIFAug increases $6.57\%$ and $3.12\%$ on Acc and AUC scores on the FF++(HQ) dataset when comparing (a) and (b). Comparing (b) with ours+ViT, we observe that DSRL significantly increases $+17.58\%$ in Acc and $+17.74\%$ in AUC, demonstrating that our DSRL fine-tuning scheme effectively adapts the pre-trained ViT model on generic real images to identify real and forged facial photos. Furthermore, from (c) and ours+ViT, FSR enhances the discriminative capability between real and fake features, leading to a $4.72\%$ and $2.29\%$ increase in Acc and AUC. This process amplifies relevant features while expressing irrelevant ones. As illustrated in (d) and ours+ViT, CIFAug yields considerable gains of $3.29\%$ in Acc and $1.92\%$ in AUC, indicating its efficacy in improving the generalization performance of deepfake detection. Finally, with all proposed components integrated, our complete method achieves the best performance, reaching $97.72\%$ in Acc and $99.31\%$ in AUC. 

\begin{table}[t]
 \setlength{\tabcolsep}{6pt}
 \renewcommand{\arraystretch}{1.2} 
  \caption{Effectiveness of the Proposed DSRL in Our Framework for Adapting ViT to the Deepfake Detection Task on FF++~\cite{FF++xception-iccv2019} Dataset. The Best Results are Highlighted in Bold Font.}
  \label{tab:5}
  \centering
  \begin{tabular}{lccc}
    \toprule
    ID   &\# Trainable Parameters &Acc (\%)  &AUC (\%) \\
    \midrule
    Adapter~\cite{Adapter2019}  &2.27M &91.29 &93.39 \\
    DSRL  &\textbf{0.28M} &\textbf{97.72} &\textbf{99.31} \\ 
  \bottomrule
  \end{tabular}
\end{table}

\begin{table}[htbp]
 \setlength{\tabcolsep}{6pt}
 \renewcommand{\arraystretch}{1.2}  
  \caption{Ablation study for effects of different rank $r$ values of DSRL on FF++~\cite{FF++xception-iccv2019} dataset. The Best Results are Highlighted in Bold Font.}
  \label{tab:12}
  \centering
  \begin{tabular}{lcccc}
    \toprule
    Metrics     &\#Params   &Acc (\%) &AUC (\%)\\
    \midrule
    $r=4$   &0.14M  &94.57 &97.72\\
    $r=16$  &0.57M  &94.12 &97.56\\ 
    $r=64$   &2.25M  &93.86 &97.31\\
    $r=8$~(ours) &0.27M  &\textbf{97.92} &\textbf{99.31} \\
  \bottomrule
  \end{tabular}
\end{table}

\vspace{0.05in}
\subsubsection{Effectiveness of DSRL}
We first extract the distinct relationships across patches in real and forged facial images using the proposed DSRL scheme to fine-tune the ViT model. To evaluate the effectiveness of this module, we replace DSRL with another parameter-efficient fine-tuning (PEFT) method, Adapter~\cite{Adapter2019}, to transfer the ViT's general ability to the deepfake detection task. 

Following its integration in Bert~\cite{Adapter2019}, we insert the Adapter behind the multi-head attention and MLP layer in each block of the encoder of ViT. Concretely, the Adapter contains a down-projection layer that reduces the dimension of feature space, a nonlinear activation function, and an up-projection layer that restores the original dimension. This process is defined as:
\begin{equation}
     x_{out}= [W^{up} \cdot \phi (W^{down} \cdot x^{T})]^{T}
\end{equation}
where $x \in \mathbb{R}^{N \times d}$ is the input of the Adapter, $W^{down} \in \mathbb{R}^{d' \times d}$ (where $d' \ll d$) represents the down-projection weight, $\phi$ is the nonlinear activation function, and $W^{up} \in \mathbb{R}^{d \times d'}$ is the up-projection weight. 

In our experiment, $d'$ and $d$ are set to 64 and 768, respectively. As shown in Table~\ref{tab:5}, our DSRL achieves remarkable performance only with $12.41\%$ of the trainable parameters of the Adapter. However, compared to the Adapter, DSRL improves Acc and AUC by $6.43\%$ and $5.92\%$, respectively. These results demonstrate that fine-tuning the pretrained ViT with our DSRL in the GenDF framework is more efficient than using an adapter for deepfake detection.

Additionally, we conduct comparative experiments to investigate the effects of different rank $r$ values of LoRA in DSRL. As illustrated in Table~\ref{tab:12}, our method with an 8-dimensional latent feature space achieves the best performance compared to 4, 16, and 64 dimensions while keeping the other parameters unchanged. These results demonstrate that the real and fake features can be effectively distinguished in an 8-dimensional space, whereas both lower and higher dimensions prove suboptimal.

\begin{table}[t]
 \setlength{\tabcolsep}{6pt}
 \renewcommand{\arraystretch}{1.2}  
  \caption{Ablation Study for Effects of Different Ways to Embed Representations on FF++~\cite{FF++xception-iccv2019} Dataset. The Best Results in Terms of Acc ($\%$) and AUC ($\%$) Are Highlighted in Bold Font.
  }
  \label{tab:8}
  \centering
  \begin{tabular}{cccccccc}
    \toprule
    ID   &Q    &K  &V  & Acc (\%)  & AUC (\%) \\
    \midrule
    (a)   &$\Delta W_{q}$ &${-}$   &${-}$ &90.14 &94.46\\
    (b)   &${-}$ &$\Delta W_{k}$   &${-}$ &91.27 &95.93\\
    (c)   &${-}$ &${-}$   &$\Delta W_{v}$ &91.43 &96.04\\
    (d)  &$\Delta W_{q}$ &$\Delta W_{k}$   &${-}$  &92.29 &96.35\\
    (e)  &{-} &$\Delta W_{k}$   &$\Delta W_{v}$  &93.14 &96.75\\  
    (f)   &$\Delta W_{q}$ &$\Delta W_{k}$   &$\Delta W_{v}$ &94.71 &97.98\\
    (g)   &$\Delta W_{q}$  &{-} &$\Delta W_{v}$  &\textbf{97.92} &\textbf{99.31}\\
  \bottomrule
  \end{tabular}
\end{table}

\begin{table}[th]
 \setlength{\tabcolsep}{6pt}
 \renewcommand{\arraystretch}{1.2} 
  \caption{Effectiveness of the proposed feature space redistribution in our framework on FF++~\cite{FF++xception-iccv2019} dataset. The best results are highlighted in bold font.}
  \label{tab:6}
  \centering
  \begin{tabular}{lcccc}
    \toprule
    ID    &\# Trainable Parameters &Acc (\%)  &AUC (\%) \\
    \midrule
    Linear Layer  &0.56M &94.00 &97.44 \\
    FSR (Ours)  &\textbf{1.5k} &\textbf{94.86} &\textbf{98.28} \\ 
  \bottomrule
  \end{tabular}
\end{table}

\subsubsection{The Effects of Different Fine-Tuning Schemes}
 In this subsection, we conduct experiments to compare the effects of different LoRA schemes employed within the self-attention modules in GenDF. As illustrated in Table~\ref{tab:8} (g), the best performance is achieved by fine-tuning $\Delta W_{q}$, $\Delta W_{v}$ while keeping the pre-trained weights $W_q$, $W_k$, and $W_v$ frozen. This strategy effectively transfers classification abilities on generic real images of $Q$ and $V$ matrices to the representation learning of real and fake facial images while preserving the basic knowledge in the $K$ matrix. However, as shown from (a) to (f) of Table~\ref{tab:8}, these fine-tuning strategies cause a significant decline in performance, weakening the discrimination between real and fake features. These findings highlight the importance of retaining the pre-trained ViT's general discrimination ability, which is crucial for deepfake detection.

\vspace{0.05in}
\subsubsection{Effectiveness of feature space redistribution}
To evaluate the effectiveness of our proposed feature space redistribution (FSR) scheme, we compare the detection performance by optimizing feature distributions with different methods. As illustrated in Table~\ref{tab:6}, we replace the FSR with a fully connected layer in the GenDF framework. The results indicate that our FSR outperforms the linear layer while requiring only $0.2\%$ of its parameters, demonstrating that our FSR efficiently optimizes the feature space for deepfake detection.

\begin{table}[t]
 \setlength{\tabcolsep}{4pt}
 \renewcommand{\arraystretch}{1.2} 
  \caption{Robustness Evaluation in Terms of AUC (\%) on FF++~\cite{FF++xception-iccv2019} Dataset.  "Avg." represents the Mean Score of These Four Perturbations. The Best Results Are Highlighted in Bold Font.}
  \label{tab:7}
  \centering
  \begin{tabular}{lccccc}
    \toprule
    Methods   &Contrast &Saturation  &Pixelate  &Blur  &Avg.\\
    \midrule
    Xception~\cite{xception}  &97.23 &98.35 &84.29  &69.12  &87.25\\
    Face-x-ray~\cite{r41-Face-x-ray}  &88.50 &97.60  &88.60 &63.80  &84.63\\
    RFM~\cite{RFM2021} &96.49 &98.62  &84.50 &58.51  &84.53\\
    RECCE~\cite{recce-cvpr2022}  &\textbf{98.87} &99.22  &90.13 &76.92  &91.29\\
    UIA-ViT~\cite{uia-vit-eccv2022}  &95.10 &96.51  &91.98 &80.24   & 92.14\\
    GenDF (Ours) &97.48 &\textbf{99.26}  &\textbf{92.51} &\textbf{85.33}  &\textbf{93.58}\\
  \bottomrule
  \end{tabular}
\end{table}

\subsection{Discussion and Analysis}
\subsubsection{Robustness Evaluation.}
The robustness of deepfake detection methods is challenged by various real-world surroundings and image-processing techniques that introduce perturbations to facial images. To evaluate the robustness of our approach, we investigate the performance under some common perturbations. Specifically, we conduct experiments on the FF++(HQ) dataset involving four types of perturbations: contrast, saturation, pixelate, and blur. As shown in Table~\ref{tab:7}, our method achieves the highest average AUC across four perturbations, outperforming the competing methods and exceeding the UIA-ViT model by 1.44\%. Notably, our approach attains state-of-the-art performance under saturation, pixelating, and blur perturbations. In particular, while other competing models usually suffer from an obvious performance decline when gaussian blur is applied to facial images, our method surpasses UIA-ViT~\cite{uia-vit-eccv2022} by $5.09\%$. Although the RECCE~\cite{recce-cvpr2022} model obtains the highest AUC score under contrast perturbation, our framework outperforms it on the remaining three perturbations, especially exceeding RECCE by 8.41\% on blur perturbation. These results indicate that our method exhibits superior robustness against various perturbations that degrade the image quality of real and fake facial images.

\begin{figure}[t]
\centering
\includegraphics[width=0.9\linewidth]{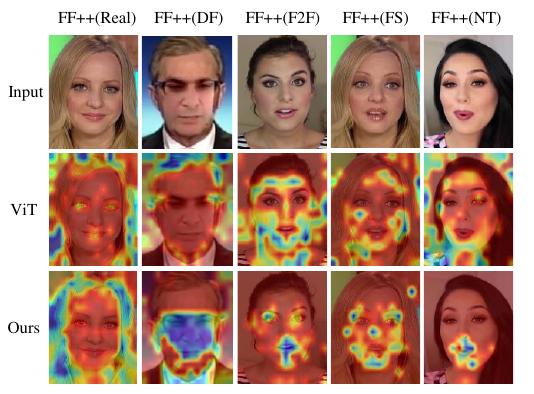}
\caption{The Grad-CAM activation map on the FF++(HQ) dataset.}
\label{CAM}
\end{figure}

\subsubsection{Patch-Discontinuity Visualization}
To analyze the decision-making mechanism of our method, we visualize Grad-CAM~\cite{2017Grad-CAM} activation maps of real and fake representations learned by the ViT backbone and our GenDF. As shown in Fig.~\ref{CAM}, GenDF precisely captures the discriminative patterns of both authentic and various forged facial images generated by various manipulation techniques, outperforming the original ViT model.

\begin{figure*}[t]
\centering
\includegraphics[width=0.98\linewidth]{figures/t-sne.pdf}
\caption{The t-SNE feature distribution visualization of real and forged facial images obtained by the baseline model (ViT) and our proposed method (Ours+ViT) under the intra-testing and the cross-testing settings. (a) ViT on the intra-testing dataset (FF++). (b) Ours+ViT on the intra-testing dataset (FF++). (c) Ours+ViT on the cross-testing dataset (DFD). (d) Ours+ViT without Class Invariant Augmentation (CIFAug) function on the cross-testing dataset (DFD).} 
\label{t-sne}
\end{figure*}

Specifically, our method captures the continuity between patches within real images, concentrating on the entire face rather than other information, such as hair and neck. Moreover, GenDF precisely attends to the boundary of the face-swapping region when confronting fake faces produced by the Deepfakes (DF) technique, aligning with our argument that the pixels at the boundary of the forged region exhibit discontinuity from the original image. Furthermore, our method successfully identifies manipulated artifacts in key facial regions such as the eyes, eyebrows, and mouth on forged images produced by Face2Face (F2F), which modifies facial expression. For fake faces generated via FaceSwap (FS) to alter identity, GenDF effectively highlights subtle artifacts around facial landmarks. Additionally, for NeuralTextures (NT)-based alterations, our method accurately locates the subtle boundaries around the mouth area caused by expression changes.
In contrast, the ViT backbone struggles to learn deepfake-specific representations. Its activations exhibit randomness when analyzing forged faces, failing to identify manipulation traces across different forgery patterns. These findings provide insights into the decision-making process of GenDF and further demonstrate its discriminative ability for real and diverse fake faces. 

\subsubsection{Analysis of Feature Distribution}
To further demonstrate the discriminative and generalized ability of our GenDF, we visualize the feature space. Specifically, we employ t-SNE~\cite{r61} to compare the feature distributions of the baseline ViT and our approach in intra-domain and cross-domain settings. Features are extracted from the layer preceding the classification head in both models. 

As shown in Fig.~\ref{t-sne}(a) and (b), the baseline ViT pretrained on generic real images fails to distinguish real and fake facial images. In contrast, integrating our GenDF into ViT enables the model to effectively learn discriminative representations, facilitating a clear distinction between real and fake images. Notably, as illustrated in Fig.~\ref{t-sne}(c), our method establishes a well-defined boundary to separate real and fake samples when evaluated on DFD, demonstrating its generalization abilities in unseen domains.
Furthermore, as shown in Fig.~\ref{t-sne}(d), Ours+ViT without CIFAug misclassifies a small number of samples. CIFAug refines the feature distributions of real and manipulated facial images along class-invariant directions to augment the original feature space. This result highlights the effectiveness of CIFAug in enhancing generalization performance for deepfake detection.

%% file: section/5_Conclusion.tex
\section{Conclusion}
In this paper, we propose GenDF, a generalizable deepfake detection framework based on the large-scale vision model ViT, designed to distinguish real and forged facial images with minimal additional trainable parameters. Our method effectively transfers the inherent recognition abilities of pre-trained ViT on generic images to the downstream deepfake detection task. First, GenDF learns deepfake-specific representations by fine-tuning ViT in a low-dimensional space. This deepfake-specific representation learning (DSRL) scheme captures the consistency between patches in real facial images and the inconsistencies across patches in forged ones. Next, we optimize the feature distributions of real and synthetic facial images by introducing a feature space redistribution (FSR) procedure, enabling the learning of more discriminative features. Additionally, we propose a feature augmentation strategy (CIFAug) that enhances generalization by actively seeking diversified representations along class-invariant directions, further improving the generalization ability on unseen real and forged facial images. Extensive experiments demonstrate that GenDF outperforms existing deepfake detection methods across various settings, including cross-domain, cross-manipulation, and intra-domain, highlighting its remarkable generalization capability in unseen scenarios.